%% file: anno2024ugc.tex
\documentclass[sigconf]{acmart}

\renewcommand\footnotetextcopyrightpermission[1]{} 
\usepackage{booktabs} 
\usepackage{url}                                                                                                       
\usepackage{bm}
\usepackage{stfloats}
\usepackage{subfigure}
\usepackage{amsmath}
\usepackage{color}
\newcommand{\figref}[1]{Fig.~\ref{#1}}
\newcommand{\tblref}[1]{Table~\ref{#1}}
\newcommand{\equref}[1]{(\ref{#1})}
\newcommand{\secref}[1]{Section \ref{#1}}

\newcommand{\argmax}{\mathop{\rm arg~max}\limits}

\makeatletter
\newcommand{\figcaption}[1]{\def\@captype{figure}\caption{#1}}
\newcommand{\tblcaption}[1]{\def\@captype{table}\caption{#1}}
\usepackage{graphicx}

\setcopyright{none}

\acmConference[SIGSPATIAL'24]{In Proceedings of the 32nd ACM SIGSPATIAL International Conference on Advances in Geographic Information Systems, October 29 - November 1, 2024}{October, 2024}{Atlanta, GA, USA}
\acmYear{1997}
\copyrightyear{2024}

\begin{document}
\title{Congestion Forecast for Trains with Railroad-Graph-based \\ Semi-Supervised Learning using Sparse Passenger Reports}



\author{Soto Anno*}
\thanks{*~This work was performed during a research internship at LY Corporation, Tokyo, Japan.}
\affiliation{%
  \institution{Tokyo Institute of Technology}
  \city{Tokyo}
  \country{Japan}
}
\email{anno@miubiq.cs.titech.ac.jp}

\author{Kota Tsubouchi}
\orcid{1234-5678-9012}
\affiliation{%
  \institution{LY Corporation}
  \city{Tokyo}
  \country{Japan}
}
\email{ktsubouc@lycorp.co.jp}

\author{Masamichi Shimosaka}
\affiliation{%
  \institution{Tokyo Institute of Technology}
  \city{Tokyo}
  \country{Japan}
}
\email{simosaka@miubiq.cs.titech.ac.jp}




\input{components/00abst.tex}

%
%



\maketitle

\input{components/01intro.tex}
  
\input{components/02related.tex}
\input{components/03pre.tex}
\input{components/04proposed.tex}


\input{components/05exp.tex}

\input{components/06discussion.tex}

\input{components/07concl.tex}

\subsection*{Acknowledgement}

This work would not have been successful without the support of many people. We could not include our entire team as co-authors, but we would like to thank them all here. We especially thank Mikiya Maruyama, Ryota Kitamura, and Rikako Takada for their cooperation in the UGC data acquisition and thoughtful discussions.

\newpage

\bibliographystyle{abbrv}
\bibliography{references}



\end{document}

%% file: components/00abst.tex
\begin{abstract}
  Forecasting rail congestion is crucial for efficient mobility in transport systems. We present rail congestion forecasting using reports from passengers collected through a transit application. Although reports from passengers have received attention from researchers, ensuring a sufficient volume of reports is challenging due to passenger's reluctance. The limited number of reports results in the sparsity of the congestion label, which can be an issue in building a stable prediction model. To address this issue, we propose a \textbf{s}emi-s\textbf{u}pe\textbf{r}vised method for \textbf{con}gestion \textbf{for}ecasting for \textbf{t}rains, or \textbf{SURCONFORT}. Our key idea is twofold: firstly, we adopt semi-supervised learning to leverage sparsely labeled data and many unlabeled data. Secondly, in order to complement the unlabeled data from nearby stations, we design a railway network-oriented graph and apply the graph to semi-supervised graph regularization. Empirical experiments with actual reporting data show that SURCONFORT improved the forecasting performance by 14.9\% over state-of-the-art methods under the label sparsity.
\end{abstract}

%% file: components/01intro.tex
\section{Introduction}

Forecasting rail congestion plays a significant role in transport systems. In fact, rail congestion leads to risks of injuring or killing commuters, for example, falling off platforms
, breaking a glass window
, or operations with doors unclosed
. Traditionally, rail congestion has been monitored using ticket gates~\cite{myojo:wit2006,sugiyama:rtri2010} and CCTV videos~\cite{tomar:dasa2022}. However, ticket gates could not quantify passengers inside individual trains. Moreover, it is well known that vision-based technique like CCTVs suffers from occlusion~\cite{tomar:dasa2022}.

\begin{figure*}[t]
    \centering
    \subfigure[Screen of the transit guidance app.]
    {
        \label{fig:app-screen}
        \begin{minipage}{0.47\hsize}
            \centering          
            \includegraphics[width=\linewidth]{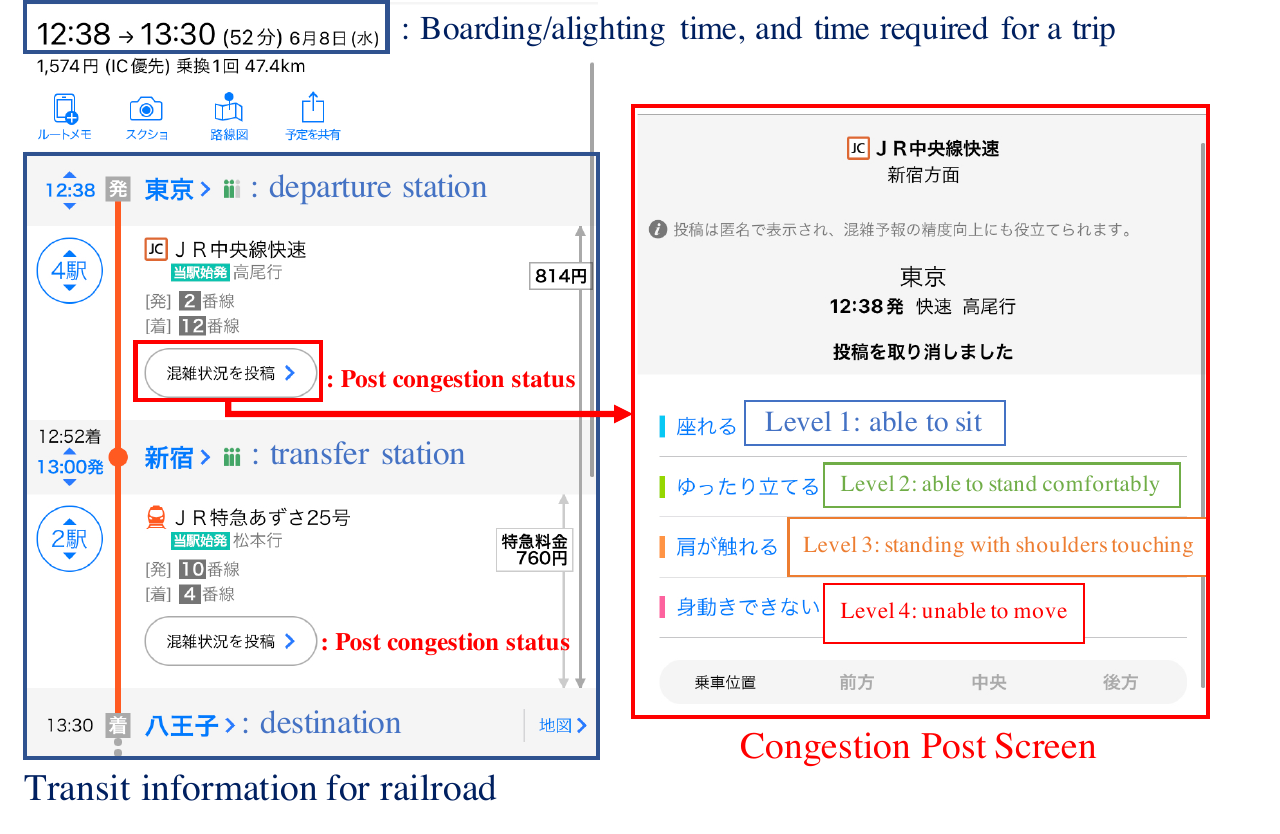}
        \end{minipage}
    }	
    \subfigure[Overview of SURCONFORT]
    {
        \label{fig:problem-overview}
        \begin{minipage}{0.47\hsize}
            \centering
            \includegraphics[width=1.0\linewidth]{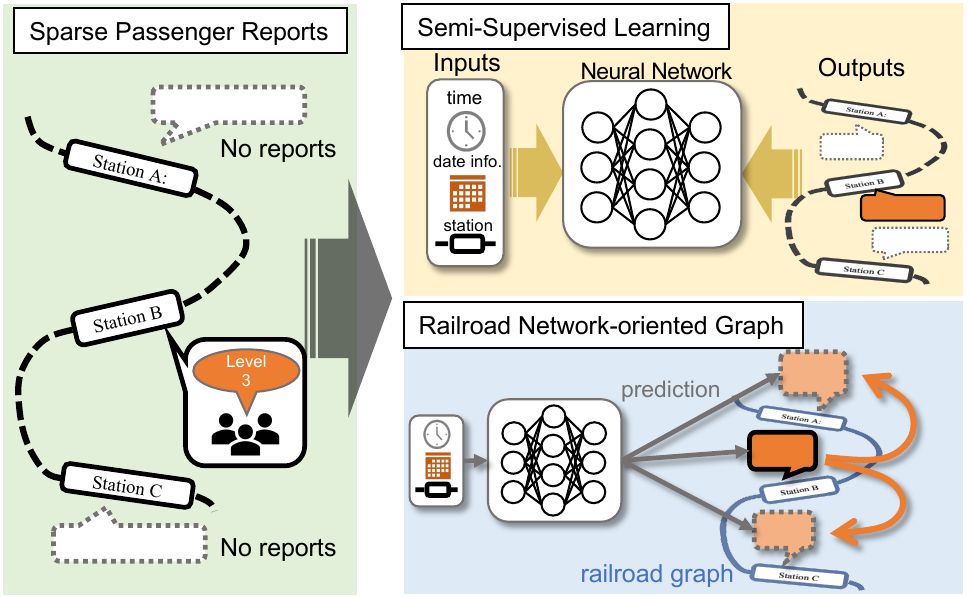}
        \end{minipage}
    }
    \caption{(a) Transit guidance screen and congestion status posting screen of 
    LY Corporation Transit Navigation. App users post the status of congestion inside the trains after boarding. (b) Overview of our problem and SURCONFORT in which the key idea is two fold: semi-supervised learning and a railroad network-oriented graph.}
    \label{fig:app-screen-and-problem-overview}
\end{figure*}

Crowdsourced information provided by passengers has received attention from researchers.
Latia et al. have revealed the efficacy of crowdsourced information to know timely situations about congestion in transportation systems~\cite{lathia:MobiQuitous2014}.
In fact, many transportation system-related applications (e.g., transit search apps such as Jorudan's Japan Transit Planner\footnote{\url{http://www.jorudan.co.jp/}}, NAVITIME\footnote{\url{https://corporate.navitime.co.jp/en/}}, LY Corporation Transit Navigation App\footnote{\url{https://transit.yahoo.co.jp/}}) have recently been collecting reports on the congestion of the trains from the passengers as shown in \figref{fig:app-screen}.
Our study aims {\bf to forecast rail congestion by leveraging the reports on the congestion submitted by rail passengers}.


However, the passenger reports are often sparse because passengers might be reluctant to submit reports on heavily crowded trains. Thus, most data, that is, railways, stations, and time slots, lack congestion labels, making it difficult to forecast the congestion stably, especially on the date and time in which there were no passenger reports in the past.

To address this issue, we propose railroad-graph-based \textbf{s}emi-s\textbf{u}pe\textbf{r}vised methodology for \textbf{con}gestion \textbf{for}ecasting of \textbf{t}rain, or \textbf{SURCONFORT}, which trains a neural network (NN) to classify the degree of congestion at a given station, date, and time. As illustrated in \figref{fig:problem-overview}, the underlying idea of SURCONFORT is twofold: (1) the adoption of semi-supervised learning (SSL) for mitigating the need for labeled data, and (2) railroad network-oriented graph for complementing predictions for unlabeled data by leveraging geospatially nearby labeled stations.

Firstly, we adopt the SSL, which has been promising in the computer vision and image classification field in recent years~\cite{guillaumin:cvpr2010,dai:iccv2013,shrivastava:eccv2012}. The SSL is a methodology that leverages both sparsely labeled data --- data annotated with congestion labels for railways, stations, and time slots --- and large amounts of unlabeled data --- data in which only covariates such as railways, stations, and time slots exist --- in order to achieve a high predictive performance compared to the model that is trained only by sparsely labeled data. Since the SSL requires a relationship between labeled and unlabeled data, the graph-based extraction of the relationship has been proposed in existing methods~\cite{ouali:2020}. However, highly scarce labeled data makes it challenging to compose graphs or leads to the propagation of label prediction errors on the graph even if the graph could be composed.

To compose an effective graph, secondly, we focus on the railroad network; in other words, our approach is to forecast congestion for unlabeled data by using labeled data in geospatially nearby stations. Cai et al. have revealed that once rail congestion has occurred, it propagates through a railroad network~\cite{cai:2022}. Consequently, nearby stations are likely to have similar congestion status, while distant stations are likely to have different congestion status. To resemble nearby stations for unlabeled data, we design a {\it railroad graph} that reflects a railroad network, wherein nodes represent stations, and edges represent stations connectivity, directionality, and spatial proximity. We then apply a graph regularization to the NN with our railroad graph in order to ensure that predictions for proximate stations on the railroad graph are similar.

Empirical experiments demonstrate that our method's superiority in tackling the data sparsity: SURCONFORT consistently outperforms state-of-the-art methods for supervised or graph-based semi-supervised learning across the varying degrees of label sparsity. Furthermore, we conduct comprehensive case studies to discuss the benefits and limitations of SURCONFORT.

The contributions of this work are as follows:

\begin{itemize}
  \item We propose SURCONFORT for forecasting train congestion by using sparse passenger reports. 
 \item We build a railway graph reflecting both line connectivity and geographical proximity to ensure that predictions for proximate stations are similar.
 \item We demonstrate the superiority of SURCONFORT over state-of-the-art methodologies by using actual reporting data collected through a transit application.
\end{itemize}



In the remaining sections that follow:
\secref{sec:related} introduces the studies related to our work.
\secref{sec:preliminaries} lays out the problem we are tackling and the baseline methods. 
\secref{sec:proposed} describes our proposed method. 
\secref{sec:exp} presents the details of our experiments. 
\secref{sec:discussion} discusses our results and how our framework can be used.
Finally, \secref{sec:conclusion} concludes our study.

%% file: components/02related.tex
\section{Related Work}
\label{sec:related}

\subsubsection*{\textbf{Railroad Congestion Estimation and Forecasting}}



Analysis of entry and exit records stored at ticket-checking gates is a mainstream approach to understanding railroad congestion. This stems from the fact that the ticket-checking data yield a more precise number of passengers than other types of data (e.g., ticket sales data, passenger number surveys, sensor-based data on vehicle spring pressure)~\cite{sugiyama:rtri2010}. Researchers extracted the origin-destination (OD) pairs from the ticket checking data and developed a method for passenger flow estimation~\cite{myojo:wit2006,sugiyama:rtri2010} and short-term forecasting~\cite{chen:chen2011}. However, as discussed in~\cite{myojo:wit2006}, OD data do not directly provide information about the rail lines and trains used by passengers.

The recent popularity of mobile device-based location histories (e.g., GPS-based mobility logs) and smartphone-based sensing has enabled analyses of congestion in a city~\cite{elhamshary:percom2018,khezerlou:trans17,jiang:kdd19,jiang:www2023,konishi:ubicomp16,anno:sigspatial2020,anno:sigspatial21}. In particular, researchers proposed a framework for predicting crowding at stations that is based on GPS-based mobility logs and transit search logs~\cite{konishi:ubicomp16,anno:sigspatial2020,anno:sigspatial21}. However, not all people near a station use the train (e.g., they may be bus passengers or visiting shops at the station), so it cannot be assured that such data accurately reflect railroad congestion. Elhamshary et al.~\cite{elhamshary:percom2018} estimated station congestion by using smartphone sensor data, including people's activity status (waiting for a train, buying a ticket, etc.) in real-time. Although compared with GPS location histories, such sensing can more accurately take into account conditions at a train station, installing sensors on each railcar would be rather costly. Moreover, some passengers might be concerned about privacy.

Apart from these approaches, Hu et al.~\cite{hu:jat2019,hu:iet2020} proposed a forecasting framework for train congestion that uses a dataset recording traffic patterns of passengers on the MRT of San Francisco. Their study is orthogonal to ours in the following respects: First, although they used complete data on rail ridership, such data are not always available due to the costs of collection. Second, they provided short-term forecasting of congestion, i.e., 5 minutes ahead, while we can forecast farther into the future, as discussed in \secref{sec:hu-vs-soto}.

\subsubsection*{\textbf{Crowdsourcing and User-Generated Content}}


Crowdsourcing and crowdsensing are being used to gather collective knowledge in various fields of data analysis ~\cite{ganti:2011,hossain:2015,pedersen:2013}. An example is an annotation task on a large-scale dataset. Typically, annotations are quite expensive to make, especially in deep learning and image recognition applications that require large amounts of data~\cite{dubey:eccv2016}. To address this issue, researchers have attempted to ensure both the quantity and quality of data by using crowdsourcing to meet this demand and, in some studies, by offering rewards~\cite{kawajiri:ubicomp14,liu:infocom2017,ge:aaai2015}.



Analysis of user-generated content (UGC) has been progressing~\cite{naab2017studies}, including on user postings on SNS (twitter, facebook, instagram, etc.)~\cite{yang2019understanding,saura2022data,mayrhofer2020user} and reviews of travel destinations and restaurants on travel reservation sites~\cite{ana2019role,ray2021user}. A typical example is Wikipedia, a user-contributed encyclopedia, which is a significant source from which search engines address users’ information needs~\cite{vincent:aaai2019}. It has been reported that UGC data is often sparse. Shao et al.~\cite{shao2022asylink} addressed the problem of data imbalances caused by the sparsity of UGC and proposed a ranking learning method to reveal the correspondence between account pairs across social platforms (i.e., User Identity Linkage; UIL). Yu et al.~\cite{yu2020category} discussed a POI category-aware methodology for recommending POIs by using user-generated sparse POI check-in data. Unlike these previous studies, this study addresses a semi-supervised learning problem setting where some data are unlabeled due to content sparsity.

\subsubsection*{\textbf{Graph-based Semi-supervised Learning}}

Semi-supervised learning is a methodology that uses a small amount of labeled data and a large amount of unlabeled data for model training~\cite{chapelle2009semi}. Obtaining large amounts of labeled data is generally costly, and semi-supervised methodologies are seen as a way to deal with this problem in deep learning, especially for image classification~\cite{guillaumin:cvpr2010,dai:iccv2013,shrivastava:eccv2012}. 

Amidst the burgeoning trend, graph-based semi-supervised learning (GSSL) methodologies have emerged, showcasing exemplary efficacy across diverse tasks, notably classification~\cite{ouali:2020}. In GSSL methods, each data point, be it labeled or unlabeled, is represented as a node in the graph, and the edge connecting each pair of nodes reflects their similarity. Pertaining to node classification, GSSL methods can be divided into methods which propagates the labels from labeled nodes to unlabeled nodes based on the assumption that nearby nodes tend to have the same labels~\cite{zhu:icml2003,wang:tkde07,xiaojin2002learning,zhou:nips03,iscen:cvpr2019}, and methods which learn node embeddings based on the assumption that nearby nodes should have similar embeddings in the descriptor space~\cite{kipf:iclr2016,tarvainen:nips2017,kearnes:jcmd2016,bui:wsdm2018,gopalan:wsdm21}.



The first group is characterized by methods such as label propagation (LP)~\cite{zhu:icml2003,wang:tkde07,xiaojin2002learning} or label spreading (LS)~\cite{zhou:nips03}, which predict unlabeled data by propagating label information for each data point on the graph. Recently, Iscen et al.~\cite{iscen:cvpr2019} have proposed Deep-SSL that made the LP applicable to deep neural networks (DNN) and have achieved state-of-the-art results in the GSSL literature. In the Deep-SSL framework, the DNN model is pre-trained only using labeled data and is expected to extract beneficial feature representations. However, the DNN tends to fail to learn from the highly scarce labeled data, which suffers from the label error feedbacks~\cite{lee:icmlw2013}.

Consequently, our attention gravitates towards the second group, epitomized by methods such as graph convolutional network (GCN) \cite{kipf:iclr2016,tarvainen:nips2017,kearnes:jcmd2016} and graph regularization~\cite{bui:wsdm2018,gopalan:wsdm21}. At a cursory glance, these techniques appear analogous, aiming to harmonize the embeddings of interconnected graph nodes. In essence, both GCN and graph regularization are recognized as distinct variations of Laplacian smoothing~\cite{kipf:iclr2016,li:aaai2018}. Due to its implementation simplicity, we employ the graph regularization, i.e., Neural Graph Machine (NGM)~\cite{bui:wsdm2018,gopalan:wsdm21}, and extend it to a railroad network-based graph regularization. 

We will further discuss the detailed disadvantage of LP-DeepSSL in \secref{sec:preliminaries} under the label sparsity and the advantages of our NGM-based approach in \secref{sec:proposed}, and show the superiority of our method over the methods in the first group.

%% file: components/03pre.tex
\section{Preliminaries}
\label{sec:preliminaries}


\subsection{Description of UGC Data}
\label{sec:ugc-data-desc}


Our goal is to forecast train congestion aggregated from user posts. Users submit the level of train congestion through a transit application after searching for train routes (as shown in \figref{fig:app-screen}). At the time of submission, the user selects the degree of congestion he/she has experienced from among four indicators (Level 1: able to sit, 2: able to stand comfortably, 3: shoulders touching, 4: unable to move). Each post consists of the last departure station, the date and time of posting, and the degree of congestion.


\subsection{Problem Formulation}

We model train congestion based on train stations' information, dates, and time periods. Let $s$ be a station, $d$ be a date, and $t$ be a time period. We define the station feature $\bm{s} \in \mathbb{R}^S$, where $S$ denotes the number of stations on a railway line. Moreover, we define the context feature of the date $d$ as $\bm{c}^{(d)} \in \mathbb{R}^{9}$, which represents the day of the week and whether it is a holiday or not. As for the time periods, we divide one day into $T$ time segments. For example, if the length of a single time period is 10 minutes, $T = 144$. We define the time feature of $t$ as $\bm{t} \in \mathbb{R}^T$. The 1-of-K encoding method is used to create these features. Finally, we define the degree of railroad congestion as the average of the user-submitted congestion levels. Its value is discretized, as $y^{(s, d, t)} \in C := \{ 0, 1, 2, 3 \}$ at station $s$ on date $d$ at time $t$.

Based on the above notation, we can express a collection of $n$ samples $X = (\bm{x}_1, ..., \bm{x}_l, \bm{x}_{l+1}, ..., \bm{x}_{n})$ with ${\bm{x}_i}^\top = [{\bm{s}_i}^\top, {\bm{c}_i}^\top, {\bm{t}_i}^\top] \in \mathcal{X}$, 
where a tuple of the station, date, and time $(s, d, t)$ is represented by a data index $i$.
The first $l$ samples ($\bm{x}_i$ for $i \in L = \{1, ..., l\}$), denoted as $X_L$, are labeled according to $Y_L = (y_1, ..., y_l)$.
The remaining $u = n - l$ samples ($\bm{x}_i$ for $i \in U = \{l+1, ..., n\}$), denoted as $X_U$, are unlabeled, since the number of submissions is very limited and no submission data exists for a certain station or date and time.

Our objective is to create an effective classifier using the labeled samples $X_L$ with $Y_L$ and unlabeled samples $X_U$. This classifier is represented by a model which takes an input example from $\mathcal{X}$ and produces a vector of class confidence scores for the degree of congestion, denoted as $f_{\bm{\theta}}: \mathcal{X} \mapsto \mathbb{R}^4$, where $\bm{\theta}$ is the model parameter. The predicted degree of congestion is the one with the maximum confidence score, 

\begin{equation}
  \hat{y}_i = \argmax_j f_{\bm{\theta}} (\bm{x}_i)_j,
\end{equation}

\noindent
where $j$ represents the $j$-th dimension of the vector.

\subsection{Prerequisite Methods}
\label{sec:prerequisite}

\begin{figure}[t]
  \begin{center}
  \includegraphics[width=\linewidth]{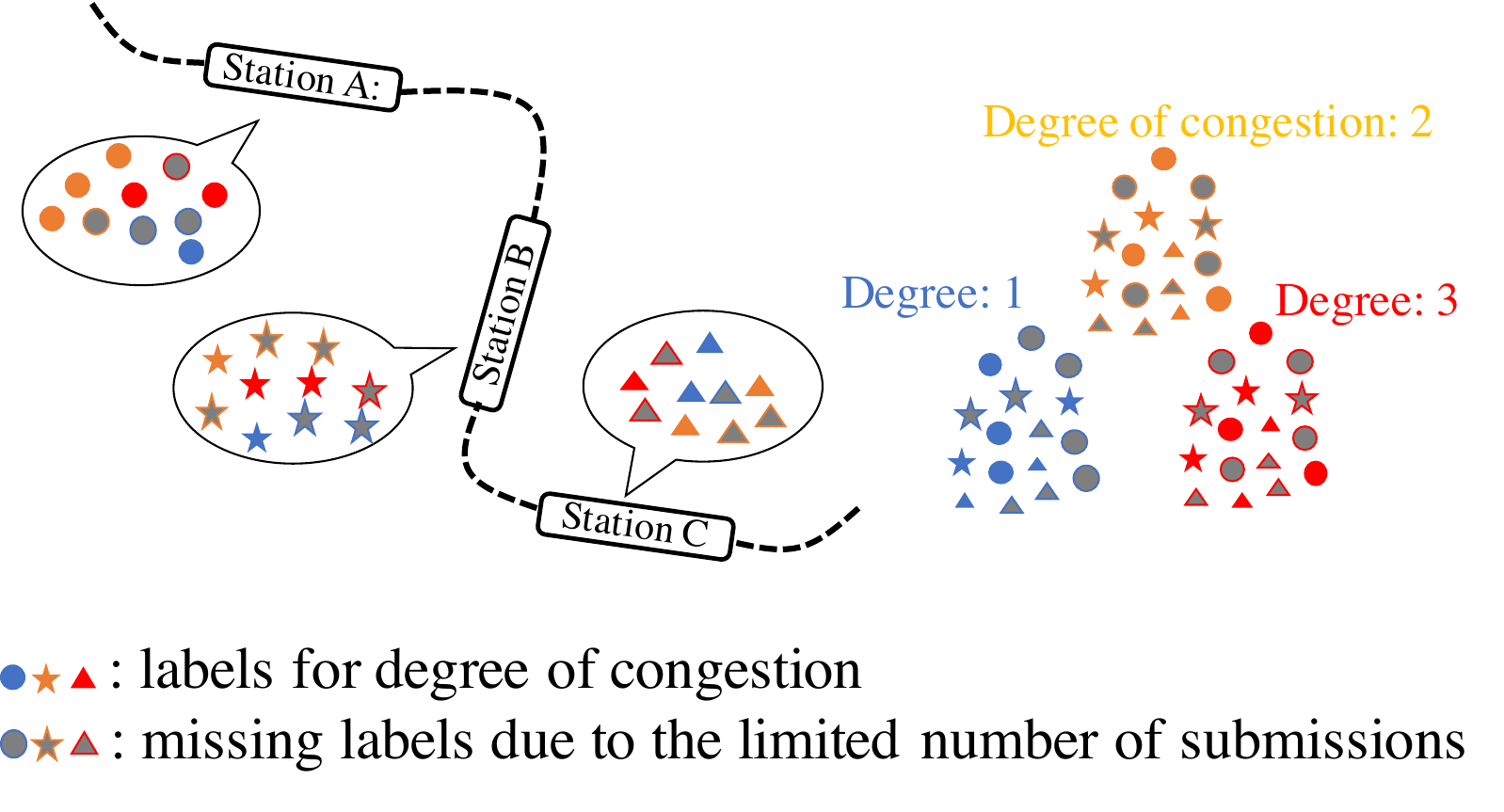}
   \caption{Conceptual illustration of our problem (left) and the descriptor space in an ideal state (right). The shape of the data point represents the sample congestion at each adjacent station (circle for station A, star for station B, triangle for station C), and the three colors represent the level of congestion at each station (blue for congestion level 1, orange for 2, red for 3). Samples missing labels in the UGC data are in gray, and the actual congestion level is reflected in the border's color. We learn the degree of congestion based on the UGC data associated with each station and build an ideal descriptor space (i.e., a cluster of descriptors for each congestion level) on the predictive model.}
  \label{fig:problem-concept}
  \end{center}
\end{figure}  

\subsubsection{Baseline (1): Simple Neural Network} 
\label{sec:baseline}
This model is constructed in a fully-supervised manner, i.e., only based on the labeled data $X_L$ and $Y_L$. The model is trained by minimizing a supervised loss $L_s$ of the form, 

\begin{equation}
  \label{eq:fully-supervised-loss}
  L_s (X_L, Y_L; \bm{\theta}) = \sum^l_{i = 1} l_s (f_{\bm{\theta}} (\bm{x}_i), y_i).
\end{equation}

\noindent
A standard choice for the loss function $l_s(\cdot, \cdot)$ in a classification task is the cross-entropy loss, which is formalized as $l_s(\bm{s}, y) = - \log \bm{s}_y $ for $\bm{s} \in \mathbb{R}^4$ and $y \in C$. 

We use four fully-connected layers with ReLU activation for the mapping $f_{\bm{\theta}}$, while any type of network can be used to form it. The feature extraction is conceptually divided into two components: {\it feature extractor } and {\it classifier}. The feature extractor is a network $\phi_{\bm{\theta}} : \mathcal{X} \mapsto \mathbb{R}^d$ that maps the input to a $d$-dimensional feature vector, the so-called {\it descriptor}. The descriptor of the $i$-th sample $\bm{v}_i$ is denoted by $\bm{v}_i = \phi_{\bm{\theta}} (\bm{x}_i)$. The classifier maps this descriptor to the class confidence score.

However, given that the size of the labeled data, i.e., $L$, is extremely small, the model's prediction performance is prone to instability. This is because a network trained on small amounts of data tends to fail to capture the characteristics behind the data in the descriptor space, such as the intra-class relationships between different stations in our case. \figref{fig:problem-concept} (right) and \figref{fig:lp-dssl-concept} (left) conceptually illustrates this situation. Ideally, data belonging to the same class should be mapped to nearby positions in the descriptor space, even between different stations (i.e., with different inputs). However, given that the model is trained with a small amount of data, the same input can only be mapped to nearby points, which results in the dispersion of the descriptors within a class. Consequently, making predictions for unknown inputs become very challenging.


\subsubsection{Baseline (2): LP-DeepSSL~\protect\cite{iscen:cvpr2019}.}

\begin{figure}[t]
  \begin{center}
  \includegraphics[width=\linewidth]{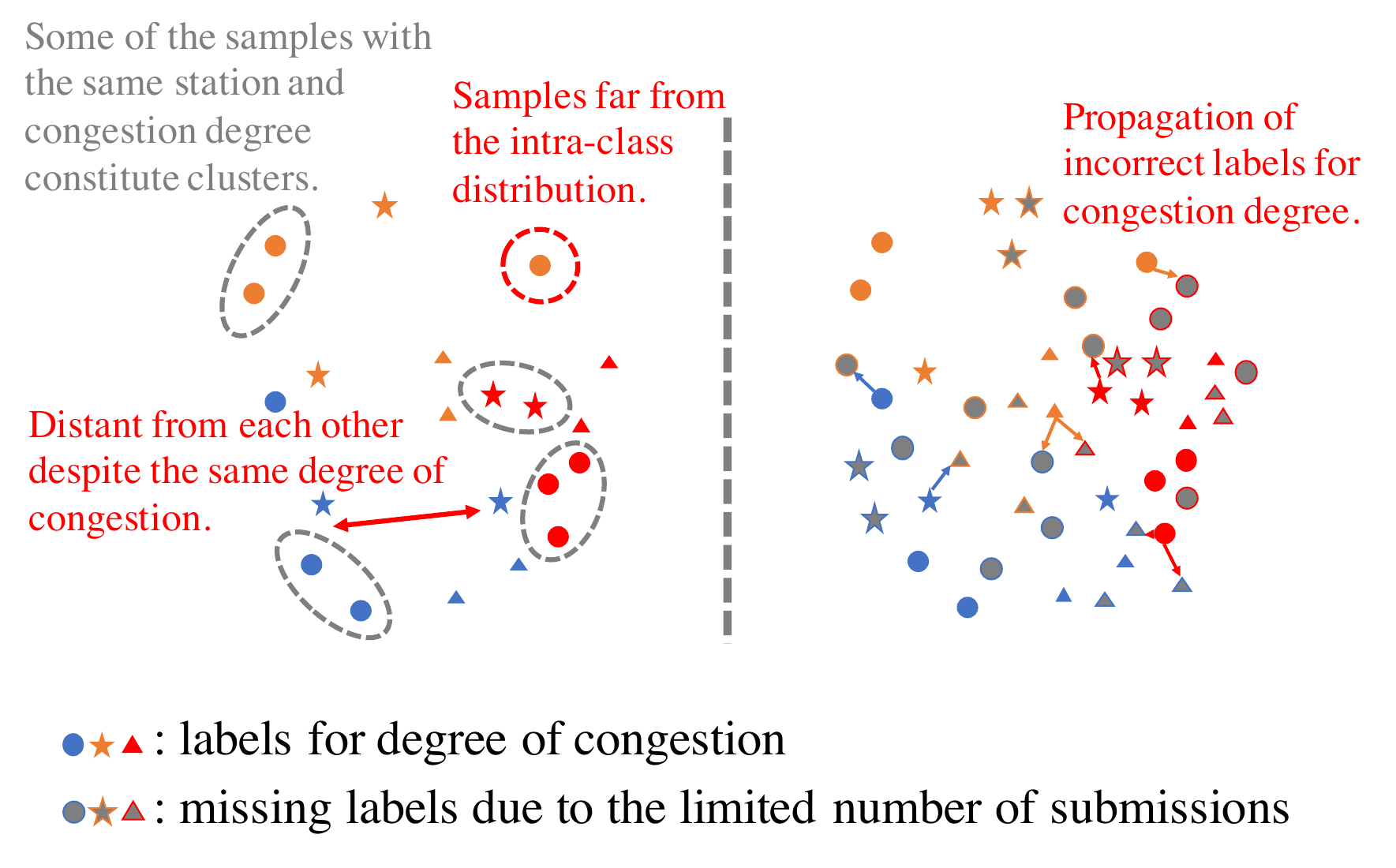}
  \caption{Conceptual illustration of descriptor space formed by fully-supervised methods (left) and LP-DeepSS~\protect\cite{iscen:cvpr2019} (right). Fully-supervised methods extract and learn limited patterns from a small number of data and thus suffer from overfitting, where some of the samples belonging to the same congestion level form clusters for each station or context. LP-DeepSSL, which is built on this basis, is prone to assigning wrong pseudo-labels (represented by unidirectional arrows) when performing label propagation on the descriptor space.}
  \label{fig:lp-dssl-concept}
  \end{center}
\end{figure}



The research community has developed a number of semi-supervised techniques to mitigate the need for labeled data. Often, the key underlying idea of the techniques is to find a feature representation that reflects the proximity between labeled and unlabeled data, and {\it graphs} --- whose each node is data and each edge is the proximity --- are typically used~\cite{ouali:2020}. 


Iscen et al.~\cite{iscen:cvpr2019} built LP-DeepSSL as an extension of label propagation~\cite{zhou:nips03} on a foundation of DNN.
This model constructs the graph representation that measures the proximities by the distance of the embeddings produced by the pre-trained model only with labeled data, $X_L$ and $Y_L$. It then performs iterative learning consisting of pseudo-labeling by label propagation and neural network training. To deal with the label propagation, a sparse {\it affinity matrix} $A := (a_{i, j}) \in \mathbb{R}^{n \times n}$ is constructed with elements,

\begin{equation}
  \label{eq:sim-matrix-in-deepssl}
  a_{i, j} = \left\{
  \begin{array}{ll}
  [{\bm{v}_i}^\top \bm{v}_j]^\gamma_+, & \mathrm{if} \ i \neq j \land \mathrm{NN}_k(\bm{v}_j) \\
  0, & \mathrm{otherwise}
\end{array}
\right.
\end{equation}

\noindent
where $\mathrm{NN}_k$ denotes the set of $k$ nearest neighbors in $X$, and $\gamma$ is a parameter following the previous work on manifold-based searches~\cite{iscen:cvpr2017}. The pseudo-labeling of the unlabeled data $X_U$, denoted as $\hat{Y}_U = (\hat{y}_{l+1}, ..., \hat{y}_n)$, is performed by label propagation with this affinity graph. The neural network is trained with the labeled and pseudo-labeled data $X_L$, $Y_L$, $X_U$, and $\hat{Y}_U$.

The affinity matrix is initially constructed by the model presented in \secref{sec:baseline}, i.e., a neural network trained in a fully-supervised manner. As discussed above, models trained in a fully-supervised manner have difficulty learning the structure of the descriptor space from a tiny amount of data. This can cause label error feedback~\cite{lee:icmlw2013} in the label propagation when pseudo-labels are mispredicted with a poor affinity matrix (as shown in \figref{fig:lp-dssl-concept}). 

%% file: components/04proposed.tex
\section{Proposed Method: SURCONFORT}
\label{sec:proposed}

\begin{figure}[t]
  \begin{center}
  \includegraphics[width=\linewidth]{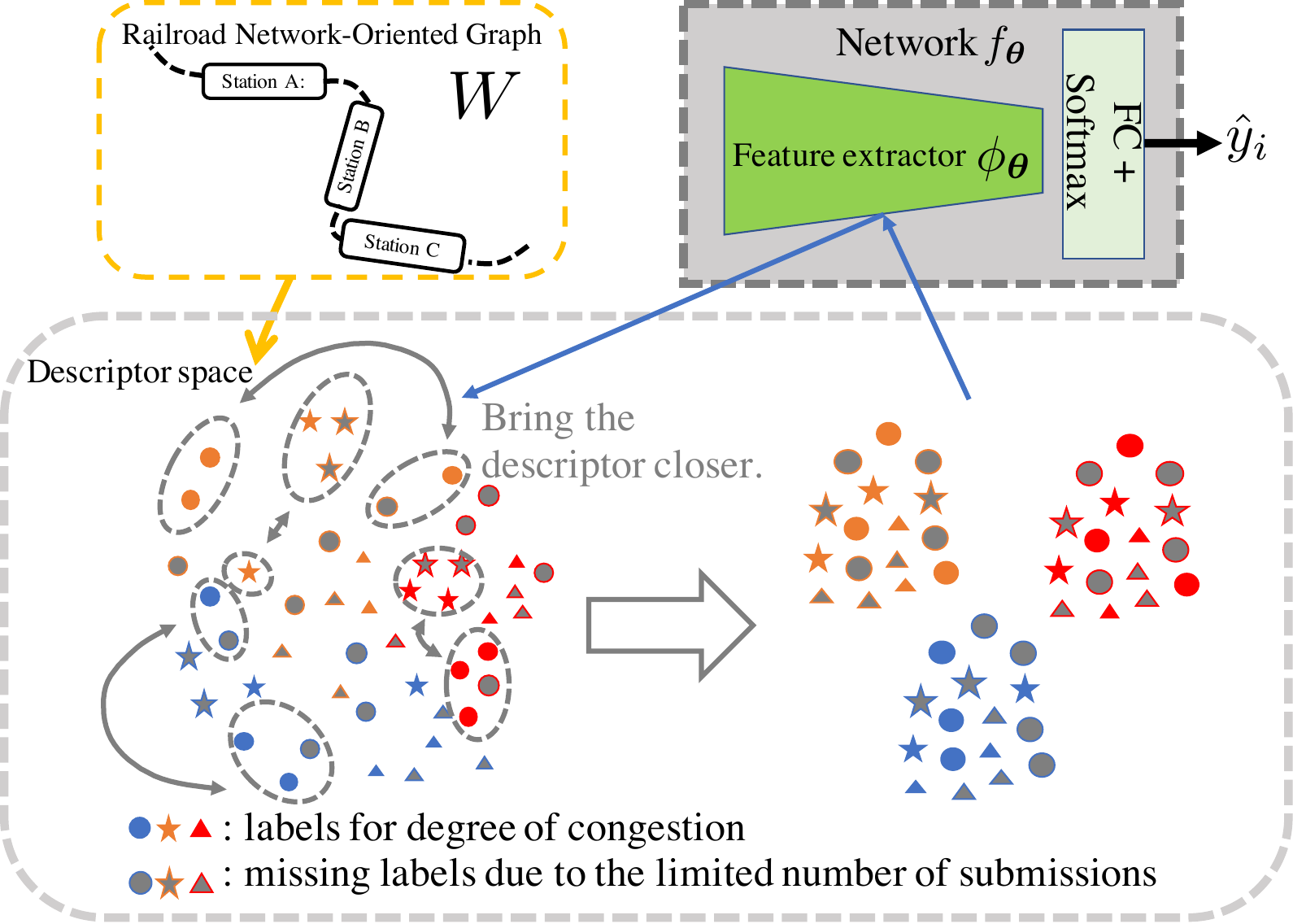}
  \caption{Conceptual illustration of SURCONFORT, which corrects the internal representation of the feature extractor by using graph regularization to create an ideal descriptor space. The graph regularization is based on a rail network-oriented graph so that the descriptors of adjacent stations are close to each other in the descriptor space. }
  \label{fig:graph-reg-concept}
  \end{center}
\end{figure}


\subsection{Basic Idea of SURCONFORT}



The discussion in the previous section indicates that the critical issue for learning to predict sparse data is forming the optimal descriptor spaces in the network. Based on our similarity assumption of congestion dynamics, the descriptor space should reflect the adjacency or spatial proximity between stations.

Our approach is to leverage the idea of graph regularization~\cite{belkin:colt2004,smola:2003,widmer:ecml2012,bui:wsdm2018} to rectify the descriptor spaces by mapping feature representations that are neighbors on the graph close to each other so the model outputs similar labels for the neighbor samples. To perform it, we use a neural graph machine (NGM)~\cite{bui:wsdm2018}, which is a class of semi-supervised learning that combines neural models and graph regularization. We construct the aforementioned railroad graph and formulate the graph regularization term based on this graph. By doing so, the trained model facilitates descriptors from different inputs (i.e., representing adjacent stations) to share similar representations if they have the same degrees of congestion. A conceptual illustration of our method is shown in \figref{fig:graph-reg-concept}.




\subsection{Neural Graph Machine with Railroad Graph}


We leverage the idea of graph regularization to deal with the dispersion issue discussed in \secref{sec:preliminaries}. The model relies on graph theory, where each node depicts a single station and each edge represents the similarity between two stations. We define the following weighted difference of two descriptors $\bm{v}_i$ and $\bm{v}_j$, as follows: 

\begin{align}
  \label{eq:rail-graph}
  \omega_\mathrm{G} (\bm{v}_i, \bm{v}_j) =  \frac{1}{2} || \bm{v}_i - \bm{v}_j ||^2_2 W_{i, j},
\end{align}

\noindent
where $W_{i, j}$ denotes a railroad network-oriented adjacency matrix reflecting the similarity between two stations $s_i$ and $s_j$ of the descriptors $\bm{v}_i$ and $\bm{v}_j$. This regularization term performs the correction of the descriptor space in a model based on the structure of the railroad network.

To define the similarity between two stations $W_{i, j}$, we focus on the heterogeneous properties of the railroad; that is, the dynamics of train congestion can propagate to adjacent stations. This phenomenon stems from the fact that the tracks of the railroad network connect stations. 
One of the simplest strategies to incorporate such an intuition into $W_{i,j}$ is to use a spatial proximity measure, such as the cosine similarity between the locations of two stations: $W_{i, j} = \cos(\bm{\sigma}_i, \bm{\sigma}_j)$, where $\bm{\sigma}_i$ denotes the spatial embedding vector for station $s_i$ (e.g., latitude and longitude), and $\cos(\bm{\sigma}_i, \bm{\sigma}_j) = \frac{\bm{\sigma}_i \cdot \bm{\sigma}_j}{||\bm{\sigma}_i || || \bm{\sigma}_j ||}$. However, this formalization does not consider actual connections on railroads or directions (up/down) of travel on a line at a station. 

Therefore, we assume that the similarity between the two stations is represented by their connections as well as spatial proximity. To incorporate such domain knowledge into the graph regularization, we define a railroad network-oriented adjacency matrix by applying a graph cut method based on the train up/down lines as follows: 

\begin{align}
  \label{eq:graph-regul}
  W_{i, j} = \left\{
  \begin{array}{ll}
  1, & \mathrm{if} \ s_i \in \phi(s_j) \ \mathrm{or} \  s_j \in \phi(s_i) \\
  1 - d/d_{\max}, & \mathrm{if} \ s_i \notin \phi(s_j), s_j \notin \phi(s_i), d < d_{\max} \\
  0, & \mathrm{otherwise}
  \end{array}
  \right.
\end{align}

\noindent 
where $d$ is the distance between station $s_i$ and $s_j$, $d_{\max}$ is a predefined maxima of $d$ to ensure the sparseness of the affinity matrix and computational efficiency, and $\phi (s)$ is the set of stations connecting to station $s$.

To perform semi-supervised learning with the graph-regularized objective, we train the model using NGM~\cite{bui:wsdm2018}. First, the fully-supervised loss function defined in \equref{eq:fully-supervised-loss} is reformulated with a graph regularization term as follows:

\begin{align}
  \label{eq:graph-regul1}
  L_\mathrm{G} (X_L, Y_L; \bm{\theta}) &= \sum^l_{i = 1} l_s (f_{\bm{\theta}} (\bm{x}_i), y_i) \nonumber \\
  & \ \ + \zeta_\mathrm{G} \sum^l_{i, j = 0} \omega_\mathrm{G} (\bm{v}_i, \bm{v}_j)
\end{align}

\noindent 
where $\zeta_\mathrm{G}$ is a hyperparameter controlling the strength of the graph regularization. This loss function can be expanded into a semi-supervised form as follows:

\begin{align}
  \label{eq:graph-regul2}
  L'_\mathrm{G} (X_L, Y_L; \bm{\theta}) &= \sum^l_{i = 1} l_s (f_{\bm{\theta}} (\bm{x}_i), y_i) \nonumber \\
  & \ \ + \zeta_\mathrm{G} \sum_{(i, j) \in \mathcal{D}_{\mathrm{LL}} } \omega_\mathrm{G} (\bm{v}_i, \bm{v}_j) \nonumber \\
  & \ \ + \zeta_\mathrm{G} \sum_{(i, j) \in \mathcal{D}_{\mathrm{LU}} } \omega_\mathrm{G} (\bm{v}_i, \bm{v}_j) \nonumber \\
  & \ \ + \zeta_\mathrm{G} \sum_{(i, j) \in \mathcal{D}_{\mathrm{UU}} } \omega_\mathrm{G} (\bm{v}_i, \bm{v}_j)
\end{align}

\noindent
where $\mathcal{D}_{\mathrm{LL}}$, $\mathcal{D}_{\mathrm{LU}}$, and $\mathcal{D}_{\mathrm{UU}}$ are sets of pairs of labeled-labeled, labeled-unlabeled, and unlabeled-unlabeled samples, respectively. This model can be trained using stochastic gradient descent.


\subsection{Conceptual Discussion}

Here, we discuss why SURCONFORT works well with railroad network-oriented graph regularization. As discussed in \secref{sec:preliminaries}, it is challenging to create an ideal descriptor space, i.e., a state where features are grouped by the degree of congestion, by using semi-supervised learning when labeled data are very scarce. This is because the graph construction in the label propagation method, i.e., Eq.~\eqref{eq:sim-matrix-in-deepssl}, does not work well with only a small amount of labeled data as it depends on the descriptor space created by the network trained in a fully-supervised manner.

Our idea is that {\it instead of creating the graph only in a data-driven manner, we incorporate domain knowledge into it for label induction} in accordance with Eq.~\equref{eq:rail-graph} and {\it rectify the descriptor space in the network} in accordance with Eq.~\equref{eq:graph-regul2}. In the first term of Eq.~\eqref{eq:graph-regul2}, the model learns the degree of congestion for the context, time of day, and station, which forms a coherent descriptor space for parts of the descriptors for the same station and date/time input. The second to fourth terms of Eq.~\eqref{eq:graph-regul2} further promote coherence in the descriptor for stations that are nearby. This cohesion is effective during forecasting because the congestion tends to propagate and its degree tends to be similar at adjacent stations.

%% file: components/05exp.tex
\section{Experiments}
\label{sec:exp}


In this section, we describe the empirical experiments on SURCONFORT that used actual data. 
First, we present the datasets and model setup. Then, we describe the methods we compared SURCONFORT against, including the state-of-the-art method for label propagation. Finally, we describe the experimental results demonstrating our prediction methodology's effectiveness and sensitivity to the strength of the graph regularization.

\subsection{Experimental Settings}

\subsubsection{\textbf{Datasets and Experimental Setups}}


We evaluated the models on a dataset consisting of actual user-submitted congestion posts. This dataset was collected using the transit search engine released by 
LY Corporation. 
The application has been downloaded over 50,000,000 times~\footnote{https://transit.yahoo.co.jp/smartphone/app/}. 
The dataset consists of six months’ worth of records from November 1st, 2020, to May 20th, 2021. As presented in \secref{sec:ugc-data-desc}, each post was made after a route search was done and while trains were running on that route. Each record contains the last departure station, the date and time of posting, and an anonymized user ID, which was deleted in the data preprocessing. We aggregated the raw posts from each station and date and time segment to calculate the average degree of congestion. We did not use any personally identifiable information in this experiment. 

We chose the JR Yamanote line for our experiments, one of Tokyo's main lines with the highest passenger numbers (31.81 million/week), implying that congestion forecasting is valuable and practical. The proposed method does not use any regional parameters. Therefore, the proposed method should work well in other areas or lines. 

We considered one day to be 24 hours and set the number of time segments $T$ to $144$ (i.e., a single time segment corresponds to 10 minutes). The period from 1:20 A.M. to 4:30 A.M. are out-of-service hours, and the data in this period were filtered out when training and testing the model. As a result, the total size of the preprocessed dataset was 2,034,779, of which 10,373 were labeled data (i.e., at least one submission existed) and 2,024,406 were unlabeled data.
We varied the amount of labeled data used for training by 10\%, 25\%, 50\%, 75\%, and 100\% of the 10,373 labeled samples to demonstrate the forecasting robustness of the proposed model in the case that labeled data are quite sparse. For the performance evaluation, we conducted a 5-fold cross-validation; i.e., we divided the dataset into five subsets and used four of them for model training and the remaining one for testing.

\subsubsection{\textbf{Model Setting}}

\label{sec:model-setting}

As for the context denoted by $\bm{c}_d$, we used the days-of-the-week and holiday-or-not features. Based on one-hot encoding, the days-of-the-week feature is a seven-dimensional vector, and the holiday-or-not feature is a two-dimensional vector. We simply concatenated these feature vectors, i.e., ${\bm{c}_d}^\top = [{\bm{c}^{(1)}_d}^\top, {\bm{c}^{(2)}_d}^\top ] \in \mathbb{R}^{9}$. For the hyper-parameter settings, we set $\delta = 0.9$, $k = 50$ and $\gamma = 3$, as in~\cite{iscen:cvpr2019}. For the graph regularization term, we set $\zeta_{\mathrm{G}} = 0.7$ in the evaluation presented in \secref{sec:results}, while we conducted a sensitivity study with this strength of regularization in \secref{sec:sensitivity}. 

\subsubsection{\textbf{Comparison Methods}}
\label{sec:comparisons}



We compare SURCONFORT with the following baselines:
\begin{itemize}
  \item \textbf{Random}: This method randomly predicts the labels.
  \item \textbf{MODE}: This method returns the most frequent labels in the training data with the same day of the week and time. If there is no such data, it returns a randomly chosen label.
  \item \textbf{SNN}: This simple neural network, presented in \secref{sec:prerequisite}, is trained in a {\bf fully-supervised manner}. The model has four fully-connected layers followed by ReLU activation in the first three layers and Softmax activation in the final layer. The number of output dimensions for all coupled layers is, in order, 128, 256, 128, and 4. We inserted batch normalization layers before each fully-connected layer except the first one.
  \item \textbf{LP} (Label Propagation)~\cite{zhu:icml2003}: A pioneering method in {\bf graph-based semi-supervised learning}, this method conducts label induction by using the so-called {\it natural graph}, defining the similarity between two samples based on the L2 distance in the input space.
  \item \textbf{LS} (Label Spreading)~\cite{zhou:nips03}: Another conventional {\bf graph-based semi-supervised learning}. This method also uses natural graphs.
  \item \textbf{LP-DSSL}~\cite{iscen:cvpr2019}: A state-of-the-art method for {\bf graph-based semi-supervised learning}, presented in \secref{sec:prerequisite}. This method uses the descriptor-based graph.
\end{itemize}

\noindent
To avoid over-fitting, we implemented early stopping in all of the models. For LP-DSSL, we trained the initial models by using fully-supervised learning for 100 epochs, selected the model with the best prediction performance from those epochs, and performed the label propagation and model re-training.

\subsubsection{\textbf{Implementation}}

We used tensorflow~\cite{tensorflow2015} to build the models listed in \secref{sec:comparisons}. For the graph regularization, we used neural structured learning~\cite{arjun:wsdm21} provided with tensorflow. For LP and LS models, we used the publicly available implementations in scikit-learn~\cite{scikit-learn}. We built the implementation of LP-DSSL on top of the publicly available Pytorch~\cite{pytorch2019} code~\footnote{https://github.com/ahmetius/LP-DSSL}. All models were optimized using Adam~\cite{kingma2014adam} with a learning rate of 0.0001.



\subsection{Experimental Results}

\subsubsection{Performance Comparison}
\label{sec:results}

\begin{table*}[t]
    \centering
    \tblcaption{Performance comparison for predicting the degree of railroad congestion. The second column shows the learning protocols, in which "stats." means a statistical method, and SL and SSL are supervised learning and semi-supervised learning, respectively.}
    \scalebox{1.1}{ 
    \begin{tabular}{l l l r  r  r  r  r } 
        \toprule[1pt]
        \multicolumn{3}{l}{} & \multicolumn{5}{c}{{\bf Ratio of labeled data (\%)}} \\ 
        \cmidrule(lr){4-8}
        Model & Protocol & Graph & $10\%$ & $25\%$ & $50\%$ & $75\%$ & $100\%$ \\
        \cmidrule(lr){1-8}
        Random & - & - & $ 24.60 \pm 0.99$ & $25.05 \pm 0.42$ & $25.81 \pm 0.48$ & $24.34 \pm 0.97$ & $25.46 \pm 1.02$ \\
        MODE & stats. & - & $ 28.03 \pm 1.29$ & $31.66 \pm 0.93$ & $37.34 \pm 0.80$ & $41.42 \pm 0.62$ & $44.61 \pm 0.51$ \\
        SNN & SL & - & $54.94 \pm 0.91$ & $56.15 \pm 1.01$ & $56.23 \pm 0.68$ & $57.42 \pm 1.80$ & $58.75 \pm 0.63$  \\ 
        LP~\cite{zhu:icml2003} & {\bf SSL} & natural & $51.99 \pm 1.37$ & $52.92 \pm 1.19$ & $54.42 \pm 1.33$ & $56.01 \pm 1.44$ & $56.84 \pm 1.34$ \\ 
        LS~\cite{zhou:nips03} & {\bf SSL} & natural & $51.99 \pm 1.37$ & $52.92 \pm 1.19$ & $54.42 \pm 1.33$ & $56.02 \pm 1.45$ & $56.83 \pm 1.34$ \\ 
        LP-DSSL~\cite{iscen:cvpr2019} (SOTA) & {\bf SSL} & descriptor & $49.38 \pm 1.36$ & $52.97 \pm 1.07$ & $55.35 \pm 0.98$ & $57.10 \pm 1.52$ & $58.88 \pm 1.08$ \\ 
        \cmidrule(lr){1-8}
        {\bf SURCONFORT} & {\bf SSL} & {\bf rail} & ${\bf 56.76 \pm 1.93}$ & ${\bf 58.08 \pm 0.95}$ & ${\bf 59.41 \pm 1.37}$ & ${\bf 60.52 \pm 1.43}$ & ${\bf 60.35 \pm 1.34}$ \\ 
        \bottomrule[1pt]
      \end{tabular}
      }
    \label{tab:overall-eval}
\end{table*}

The experimental results are presented in \tblref{tab:overall-eval}. We used classification accuracy as the evaluation criterion. 
SURCONFORT performed the best out of the graph-based approaches at all the percentages of the labeled data. 

We observed attenuation in prediction performance for all models as the number of labeled data used for training decreased. This indicates that the label sparsity of UGC data is a severe problem that makes prediction difficult. Nevertheless, the proposed method succeeded in minimizing the reduction in prediction performance. Specifically, SURCONFORT improved the forecasting performance by 9.2 \% compared to LP and LS, 14.9 \% compared to LP-DSSL, and 3.3 \% compared to SNN when the model was trained using 10\% of the data. Note that the proposed method achieved the highest accuracy in all rounds of 5-fold cross-validation. This indicates that the performance improvement was significant. On the other hand, LP-DSSL performed worst in machine learning-based methods, excluding Random and MODE, when the labeled data amounted to 10\%, whereas it outperformed LP, LS, and SNN when all of the data was labeled. This implicitly demonstrates the conceptual challenge of LP-DSSL when the number of label data is significantly low, as described in \secref{sec:prerequisite}. This issue is discussed in detail in \secref{sec:surconfort-vs-lpdssl}. From these results, it can be concluded that SURCONFORT is effective at forecasting congestion on trains, even in situations where labeled data are incredibly scarce.

\subsubsection{Sensitivity Analysis of Graph Regularization}
\label{sec:sensitivity}

\begin{figure}[h]
    \centering
    \includegraphics[width=\linewidth,clip]{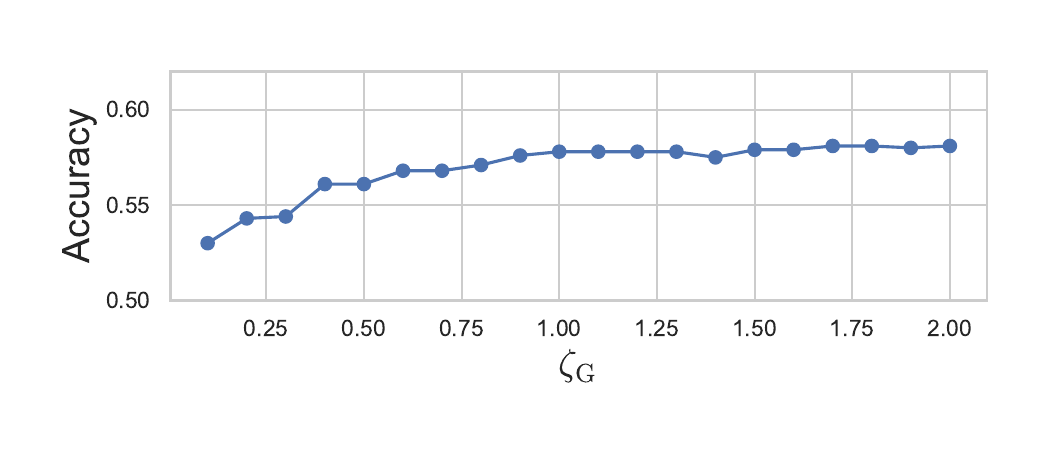}
    \figcaption{Prediction performance w.r.t. $\zeta_{\mathrm{G}}$.}
    \label{fig:sa-graph-reg}
\end{figure}

The findings from the sensitivity analysis regarding the potency of graph regularization are illustrated in \figref{fig:sa-graph-reg}. In this context, we scrutinized the sensitivity under the condition that the model was trained with merely 10\% of labeled data. When $\zeta_{\mathrm{G}} = 0$, the methodology coincides with the SNN. It is noticeable that enhancements in predictive efficacy plateau around $\zeta_{\mathrm{G}} = 1.0$, while it diminishes as the regularization becomes more lenient. This observation intimates that in situations with limited labeled data, rigorous application of regularization might lead to performance on par with scenarios having abundant labels.


\subsubsection{Ablation Study}
\label{subsubsec:ablation}

SURCONFORT is characterized by (1) the adoption of SSL and (2) the railroad graph as presented in \secref{sec:proposed}. To analyze the importance of each contribution, we assessed the forecasting performance using the following variants of SURCONFORT: (1) without railroad graph, which is the same as the original NGM~\cite{bui:wsdm2018} trained by using a natural graph, and (2) without both the SSL and railroad graph, which is the same as the SNN discussed in \secref{sec:prerequisite}. 

\tblref{tab:ablation-eval} summarizes the substantial contribution of the adoption of SSL and railroad graph towards performance enhancement. SURCONFORT achieved a performance improvement of up to 3.3 \% compared to SNN (i.e., without both the SSL and railroad graph) and up to 1.7\% compared to NGM (i.e., without the railroad graph). These results show the effectiveness of the combinations of SSL and railroad graphs in the proposed method.

\begin{table*}[t]
    \centering
    \tblcaption{Performance comparison for ablation study. SSL stands for semi-supervised learning.}
    \scalebox{1.15}{ 
    \begin{tabular}{l c c r  r  r  r  r } 
        \toprule[1pt]
        \multicolumn{3}{l}{} & \multicolumn{5}{c}{{\bf Ratio of labeled data (\%)}} \\
        \cmidrule(lr){4-8}
        Model & SSL & railroad graph & $10\%$ & $25\%$ & $50\%$ & $75\%$ & $100\%$ \\
        \cmidrule(lr){1-8}
        {\bf SURCONFORT} & $\checkmark$ & $\checkmark$ & ${\bf 56.76 \pm 1.93}$ & ${\bf 58.08 \pm 0.95}$ & ${\bf 59.41 \pm 1.37}$ & ${\bf 60.52 \pm 1.43}$ & ${\bf 60.35 \pm 1.34}$ \\ 
        NGM~\cite{bui:wsdm2018} & $\checkmark$ & - & $55.96 \pm 0.15$ & $57.69 \pm 1.15$ & $58.67 \pm 0.47$ & $59.49 \pm 0.72$ & $60.09 \pm 0.89$ \\ 
        SNN & - & - & $54.94 \pm 0.91$ & $56.15 \pm 1.01$ & $56.23 \pm 0.68$ & $57.42 \pm 1.80$ & $58.75 \pm 0.63$ \\ 
        \bottomrule[1pt]
      \end{tabular}
      }
    \label{tab:ablation-eval}
\end{table*}

\subsection{Case Studies}
\subsubsection{Performance Evaluation for Each Station}

\begin{figure*}[t]
  \begin{center}
  \includegraphics[width=1.0\linewidth]{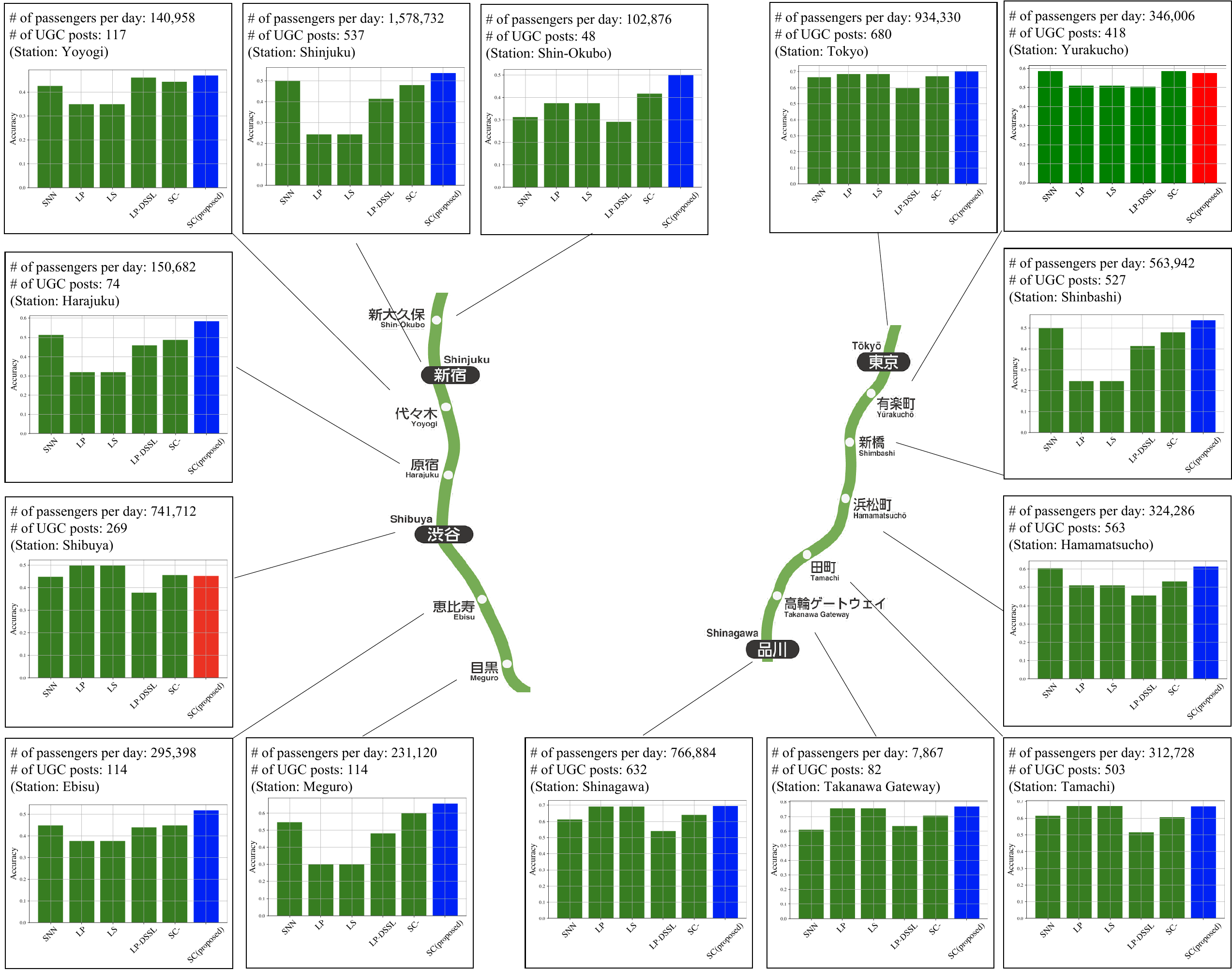}
  \caption{Performance evaluation for each station in Yamanote Line. This figure compares the number of daily users at each station, the number of passenger reposts (denoted as {\it UGC posts}) in the dataset, and the model's prediction accuracy for each station. The bar color means the proposed method outperforms the baselines if it is blue; A red color represents that the proposed method is inferior to the baselines.}
  \label{fig:eval-for-station-tokyo}
  \end{center}
\end{figure*}

\figref{fig:eval-for-station-tokyo} shows the results of the prediction performance evaluation at each station on the JR Yamanote Line. The figure compares the number of daily users at each station, the number of passenger reports in the dataset, and the model's prediction accuracy for each station.

On the left side of the figure (lines including Shinjuku and Shibuya), the proposed method outperformed other methods, except at Shibuya. The stations where it excelled tend to have fewer passenger reports, leading to label sparsity, compared to the stations on the right side of the figure. These observations suggest that the modeling approach in the proposed method, which considers the similarity of congestion levels among neighboring stations, was particularly effective in conditions of label sparsity.


However, there were instances where the comparison method performed best, notably at Shibuya and Yurakucho.

Shibuya Station is one of the largest stations with ten train lines. In contrast, adjacent stations, such as Ebisu and Harajuku, have a significantly lower daily user count and fewer lines – four and one, respectively. Shinjuku Station shares characteristics with Shibuya, but there's a notable difference in prediction performance. This discrepancy can likely be attributed to two factors: First, descriptors at Shibuya Station might have been adjusted closer to its neighboring stations despite the considerable differences in station characteristics. Second, the number of passenger reports at Shibuya is roughly half that of Shinjuku, suggesting the signal from the correct label might not have been strong enough to maintain prediction accuracy.



The number of passenger reports at Yurakucho Station is relatively assured, although the station characteristics, such as the number of passengers, differ somewhat from those of the connecting stations. The forecasting performance is also competitive with the best comparison method. From these points, we conclude that the performance degradation of the proposed method at Yurakucho is coincidental and does not pose a problem in practical applications.




\subsubsection{Visualization of Forecasting}


\begin{figure}[h]
    \centering
    \subfigure[Shinjuku, Weekday]
    {
        \label{fig:shinjuku-weekday}
        \begin{minipage}{1.0\hsize}
            \centering          
            \includegraphics[width=\linewidth]{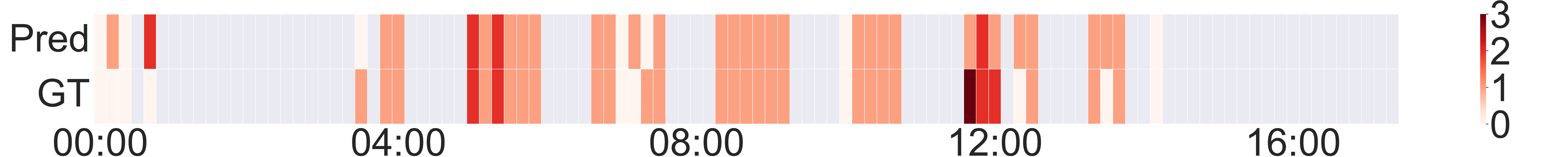}
        \end{minipage}
    }	
    \subfigure[Shinjuku, Holiday]
    {
        \label{fig:shinjuku-holiday}
        \begin{minipage}{1.0\hsize}
            \centering
            \includegraphics[width=1.0\linewidth]{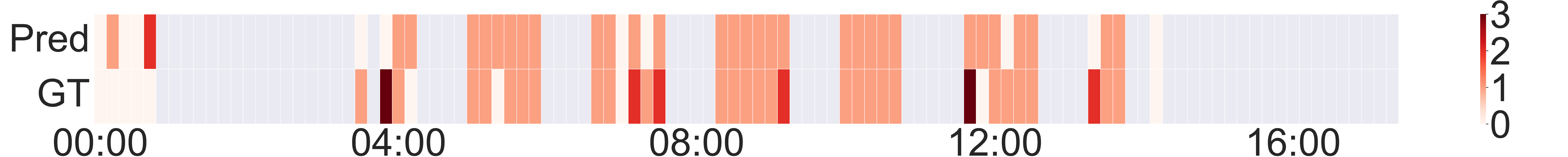}
        \end{minipage}
    }
    \subfigure[Ebisu, Weekday]
    {
        \label{fig:ebisu-weekday}
        \begin{minipage}{1.0\hsize}
            \centering
            \includegraphics[width=1.0\linewidth]{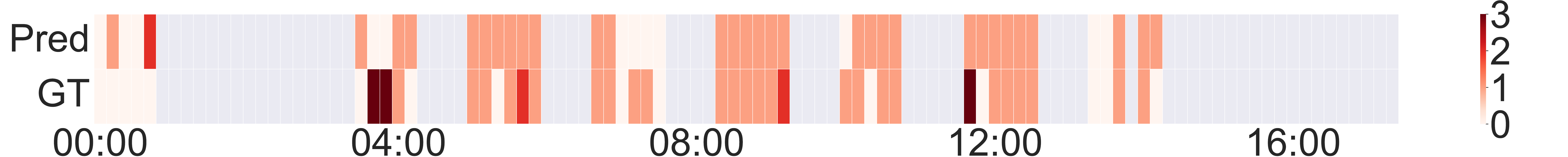}
        \end{minipage}
    }
    \subfigure[Ebisu, Holiday]
    {
        \label{fig:ebisu-holiday}
        \begin{minipage}{1.0\hsize}
            \centering
            \includegraphics[width=1.0\linewidth]{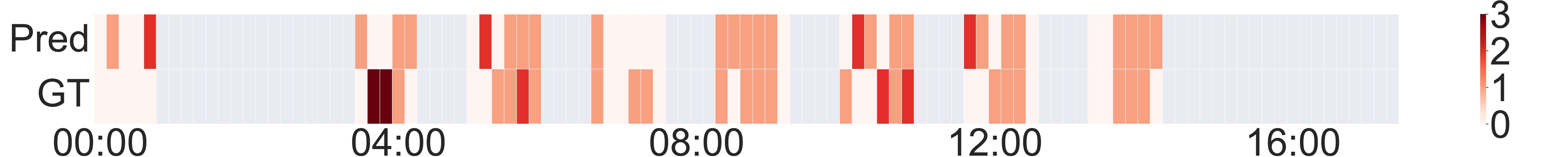}
        \end{minipage}
    }
    \caption{Visualization of forecasting by SURCONFORT at Shinjuku and Ebisu station. One colored instance represents the predicted or ground-truth congestion label in a 10 minutes time period.}
    \label{fig:case-study-vis}
\end{figure}

This section presents a case study on congestion forecasting using SURCONFORT. Figure \figref{fig:case-study-vis} visualizes the prediction results for Shinjuku Station on both a weekday (\figref{fig:shinjuku-weekday}) and a holiday (\figref{fig:shinjuku-holiday}), as well as for Ebisu Station on a weekday (\figref{fig:ebisu-weekday}) and a holiday (\figref{fig:ebisu-holiday}). Note that the gray areas represent unlabeled samples in the test data, corresponding to times with no passenger reports.

Shinjuku holds the distinction of being the station with the highest daily user count in Japan. Accordingly, it boasts a relatively large number of passenger reports. This abundance of data ensures that the forecasts are both accurate and stable, as exemplified during the periods 5:00–6:00, 8:20–9:20, and 10:00–10:50 in \figref{fig:shinjuku-weekday}, and 10:00–10:50 in \figref{fig:shinjuku-holiday}.

Conversely, there are intervals where SURCONFORT misclassified the congestion levels at Ebisu Station, specifically during 5:00–6:00 and 8:20–9:20 as shown in \figref{fig:ebisu-weekday}, and 5:00–6:00 in \figref{fig:ebisu-holiday}. This can likely be attributed to the fact that Ebisu Station accommodates fewer users compared to Shinjuku Station, with its passenger reports being roughly one-fifth of Shinjuku's. Despite the preceding section confirming the high quantitative performance of SURCONFORT, there may be instances of limited accuracy, particularly when faced with significant data sparsity. Furthermore, a consistent overestimation of congestion is observed around 4:00 in the morning at both stations. During these early hours, the passenger count is low and the passenger reports are sparse, factors which could contribute to the prediction inaccuracies.

%% file: components/06discussion.tex
\section{Discussion}
\label{sec:discussion}

\subsection{Analysis of Experimental Results}
\label{sec:surconfort-vs-lpdssl}


As highlighted in \secref{sec:results}, LP-DSSL, which employs label propagation and model retraining based on the SNN, did not achieve results comparable to SURCONFORT, especially in situations with sparse label data. This subpar performance of LP-DSSL can be attributed to the descriptor space, learned within the network, exhibiting a compromised intra-class distribution, thereby impairing effective pseudo-labeling.

In the findings of the ablation study presented in \tblref{tab:ablation-eval} of \secref{subsubsec:ablation}, we observed a progressive enhancement in prediction performance in the sequence: SNN, NGM, and finally, SURCONFORT. Such outcomes reinforce the effectiveness of both adopting semi-supervised learning and tweaking the descriptor space via graph regularization. The results also underscore the benefits of incorporating knowledge regarding station adjacencies. 
Although the margin of performance improvement between NGM and SURCONFORT may appear small, it can be rationalized as follows: NGM, when regularized by a natural graph that reflects L2 similarities of input vectors (which incorporate station one-hot encodings as edges), inherently gets partially regularized by the proximity of identical stations. In essence, NGM can be seen as a specific variant of SURCONFORT. Consequently, NGM achieves a performance that rivals that of SURCONFORT.



\subsubsection*{\textbf{Forecasting by current observations vs. forecasting by contexts}}
\label{sec:hu-vs-soto}


As discussed in \secref{sec:related}, train congestion forecasts can be broadly categorized into two types: those that provide context-based forecasts, as our method does, and those that deliver time-series forecasts rooted in current observations, such as those described in ~\cite{hu:jat2019,hu:iet2020}. Time-series forecasting operates on the premise that present observations correlate with congestion in the near future. However, leveraging this approach for long-term predictions—where such immediate correlations might not hold, such as using today's observations to forecast congestion three days ahead—is challenging. In contrast, our forecasting method's strength lies in its independence from current observations. Instead, it relies solely on the context, specifically the day of the week and the time of day. This ensures our forecasting capability isn't constrained to just the immediate future.


\subsection{Application using SURCONFORT}
The primary objective of this study was to enhance the efficacy of 
the congestion forecasting system deployed in Yahoo! JAPAN’s transit search application\footnote{https://blog-transit.yahoo.co.jp/congestion/}.

\textit{Practical Impact of Real-world Application.}
This system serves a crucial function, empowering users to navigate and avoid congested areas while providing organizations the data needed to mitigate overcrowding risks. Its value was particularly pronounced during the COVID-19 pandemic in Japan. The congestion forecasting system assesses transportation system congestion levels, proactively offering insights to a vast user community.

\textit{Milestones for Deployment.}
This research introduced a novel framework designed to deliver more \textit{authentic} congestion predictions. As it stands, the system utilizes a forecasting model proposed by Konishi et al.~\cite{konishi:ubicomp16}, drawing on transit search logs to estimate train passenger counts. However, the intricate nature of transit search behaviors—such as repeated searches by one user or searches that don't culminate in actual train boarding—casts doubt on the veracity of these logs as true reflections of passenger volumes. Given this perspective, forecasts rooted in the passenger reports can arguably offer a more accurate representation. In light of these findings, there's an ongoing initiative to supersede the existing forecasting mechanism with the newly proposed method.


\subsection{Applicability to other problems}



SURCONFORT offers utility across a range of problems where the graph structure of data serves as domain knowledge, especially in scenarios marked by limited labeled data. Our research zoomed in on the tangible challenge of train congestion, suggesting a strategy that leverages crowdsourced data coupled with railroad network-oriented graph regularization. However, the broader challenge of sparse crowdsourced data echoes across various domains beyond railway congestion.


For instance, the domain of crowd-sourced weather data faces similar challenges ~\cite{chakraborty:aas2020}. Such data, inherently more granular than satellite-derived information, carries "hyper-local" nuances. Nevertheless, given the modest quality of smartphone sensors, the weather data acquired is often riddled with noise. Additionally, there might not always be a sufficiently large dataset of high quality. In such contexts, We believe that SURCONFORT could serve as a robust tool for weather forecasting. By recognizing the geographic proximity and intrinsic physical laws, it becomes feasible to construct graphs between forecasting nodes, enhancing prediction accuracy.


\subsection{Limitations and Future Work of SURCONFORT}

We are aware that the passenger reports may reflect their subjectivity, which could be data noise in model building. Noise issues in crowdsourcing results, such as report data, have been studied extensively~\cite{hsueh2009data}, and future research could combine such techniques with SURCONFORT to achieve predictions without passenger subjectivity.

This study did not account for stations that serve multiple lines. Furthermore, factors such as the ease of transferring between lines — measured by the distance between platforms — and the availability of alternative transportation options near stations should also influence the concept of "station proximity." In future research, we intend to incorporate these additional details into our proximity information to further enhance prediction performance.

%% file: components/07concl.tex
\section{Conclusion}
\label{sec:conclusion}


We presented SURCONFORT, a novel approach to forecasting rail congestion. This method leveraged congestion reports provided by passengers. In order to complement the unlabeled data by using sparse labeled data in nearby stations, we proposed a railway network-oriented graph and applied it to the semi-supervised regularization. Experimental results using actual report data showed the efficacy of the SURCONFORT. Compared to the state-of-the-art graph-based semi-supervised learning method, our model improved forecasting performance by 14.9\% under the label sparsity. 

Future work will address the passenger subjectivity in their reports by combining SURCONFORT with the data selection technique for crowdsourcing such as~\cite{hsueh2009data}. The distance between platforms and the availability of alternative transportation options should also be considered for constructing the robust railroad graph. We plan to incorporate such side information into our proximity metric in the railroad graph.